%% file: main_FusMinCq.tex
\documentclass[a4paper]{article}

\usepackage{graphicx}
\usepackage{fancyhdr}
\usepackage{fancybox}
\usepackage{natbib}
\usepackage{hyperref}
\usepackage{xcolor}

\input{entete}

\newcommand{\version}{V 1.0}

\begin{document}

\lhead{E. Morvant, A. Habrard, S. Ayache} 
\rhead{PAC-Bayesian Majority Vote for Late Classifier Fusion}
\rfoot[Technical Report \version]{\thepage} 
\cfoot{} 
\lfoot[\thepage]{Technical Report \version}

\renewcommand{\headrulewidth}{0.4pt}  
\renewcommand{\footrulewidth}{0.4pt}

\title{PAC-Bayesian Majority Vote for Late Classifier Fusion\thanks{This work was supported in part by the french project VideoSense ANR-09-CORD-026  of the ANR in part by the IST Programme of the European Community, under the PASCAL2 Network of Excellence, IST-2007-216886. This publication only reflects the authors' views.}}

 \author{
Emilie Morvant
\and St{\'e}phane Ayache
\and Aix-Marseille Univ., LIF-QARMA, CNRS, UMR 7279, F-13013, Marseille, France\\ \textit{\{firstname.name\}@lif.univ-mrs.fr}
\and  Amaury Habrard
\and Univ. of St-Etienne, Lab. Hubert Curien, CNRS, UMR 5516, F-42000, St-Etienne, France\\ \textit{amaury.habrard@univ-st-etienne.fr}
}

\maketitle

\input{abstract}


\input{introduction}

\input{notations}

\input{contrib}

\input{experiments}

\input{conclusion}

\bibliography{biblio}
\bibliographystyle{apalike}

\end{document}

%% file: entete.tex
\usepackage{bbm}
\usepackage{graphicx}
\usepackage{booktabs}
\usepackage{multirow}

\usepackage{amsmath}

\usepackage[tight,footnotesize]{subfigure}

\usepackage{url}
\usepackage{amssymb}
\usepackage{times}
\usepackage{algorithm}
\usepackage{algorithmic}
\usepackage{enumerate}

\usepackage{color}

\newcommand{\xbf}{\ensuremath{\mathbf{x}}}
\newcommand{\Dcal}{\ensuremath{\mathcal{D}}}

\newcommand{\Qcal}{\ensuremath{\mathcal{Q}}}
\newcommand{\Wcal}{\ensuremath{\mathcal{Q}}}
\newcommand{\Qbf}{\ensuremath{\mathbf{Q}}}

\newcommand{\Mcal}{\ensuremath{\mathcal{M}}}
\newcommand{\Mbf}{\ensuremath{\mathbf{M}}}
\newcommand{\mbf}{\ensuremath{\mathbf{m}}}
\newcommand{\Abf}{\ensuremath{\mathbf{A}}}
\newcommand{\Hcal}{\ensuremath{\mathcal{H}}}
\newcommand{\R}{\ensuremath{\mathbb{R}}}

\newcommand{\PR}{\operatorname{\mathbb{P}}}

\newcommand{\Var}{\operatorname{\mathbf{Var}}}
\newcommand{\sign}{\operatorname{sign}}
\newcommand{\argmin}{\operatorname{argmin}}
\newcommand{\argmax}{\operatorname{argmax}}

\newcommand{\E}{\operatorname{\mathbb{E}}}
\newcommand{\x}{\times}

\newtheorem{definition}{Definition}

\newtheorem{theorem}{Theorem}

\makeatletter
\def\captionof#1#2{{\def\@captype{#1}#2}}
\makeatother

%% file: abstract.tex
\begin{abstract}
A lot of attention has been devoted to multimedia indexing over the past few years.
In the literature, we often consider two kinds of  fusion schemes: The {\it early fusion} and the {\it late fusion}.
In this paper we focus on late classifier fusion, where one combines the scores of each modality at the decision level.
To tackle this problem, we investigate a recent and elegant well-founded quadratic program named MinCq coming from the Machine Learning PAC-Bayes theory.
MinCq looks for the weighted combination, over a set of real-valued functions seen as voters, leading to the lowest misclassification rate, while making use of the voters' diversity.
We provide evidence that this method is naturally adapted to late fusion procedure.
We propose an extension of MinCq by adding an order-preserving pairwise loss for ranking, helping to improve Mean Averaged Precision measure.
We confirm the good behavior  of the MinCq-based fusion approaches with experiments on a real image benchmark.
\end{abstract}

%% file: introduction.tex
\section{Introduction}

Combining multimodal information is an important issue in Multimedia
and a lot of 
research effort has been dedicated to this problem (see
\cite{AtreyHEK10} for a survey). 
Indeed, the fusion of multimodal inputs can bring
complementary information, from various  sources, useful
  for improving
the quality of any multimedia analysis method 
such as for semantic concept detection, audio-visual event detection,
object tracking, etc. 

The different modalities correspond generally to a relevant set of
features that can be grouped into different views. 
For example,  classical visual or textual features commonly used in
multimedia are based on 
TF-IDF, bag of words, texture, color, SIFT, spatio-temporal
descriptors, etc.
Once these features have been extracted, another step consists in using
machine learning methods in order to build classifiers able to
discriminate a given concept. 

Two main schemes are generally considered \cite{Early-Late-ACMMultimedia05}. 
In the {\it early fusion} approach,  all the available
data/features are merged into one feature vector before the learning and
classification steps. This can be seen as a unimodal classification. 
However, this kind of approach has to deal with heterogeneous data or
features which are sometimes difficult to combine. 
The {\it late fusion} model  works at the decision level by
combining the prediction scores available for each
modality. This is usually called
multimodal classification or classifier fusion. 
 {\it Late fusion} may not always outperform 
unimodal classification. 
Especially when one modality provides
significantly better results than others or when 
one has to deal with imbalanced input features.
However, {\it late fusion} scheme tends to give 
better results for learning semantic concepts in case of multimodal
video~\cite{Early-Late-ACMMultimedia05}. 
Several methods based on a fixed decision rule have been proposed  for combining classifiers such as max, min, product, sum,  etc \cite{kittler98b}.
Other approaches, often referred to as {\it stacking} \cite{Wolpert92}, need of an extra learning step.

In this paper, we address the problem of {\it late} multimodal fusion at the decision level with stacking. Let $h_i$ be the classifier that gives the score
associated with the $i^{th}$ modality for any instance $\xbf$. 
A classical method consists in looking for a weighted linear combination of the different scores,
$$
H(\xbf) = \sum_{i=1}^n q_i h_i(\xbf),
$$
where $q_i$ represents the weight associated with $h_i$.
It is usually required  that $0 \leq  q_i \leq  1$ and $\sum_{i=1}^n  q_i = 1$. 
This linear weighting scheme can be seen as a majority vote.
This approach is widely used because of its robustness, simplicity and scalability due to small computational costs \cite{AtreyHEK10}.
It is also more appropriate when there exist dependencies between
the views \cite{WuCCS04}. 
An important issue is then to find an optimal way to combine the 
scores. 
One solution is to use machine learning methods to assess the weights \cite{AtreyHEK10}. 
From a machine learning standpoint considering a 
 set of classifiers with a high diversity is generally a
desirable property \cite{Dietterich00}. 
One illustration  is given by the algorithm
AdaBoost \cite{FreundS96}, frequently used as a multimodal fusion
method. AdaBoost  weights the
classifiers according to different distributions of the training
data, introducing some diversity, but requires at least {\it weak  classifiers} to perform well. 
Another recent approach based on the portfolio theory \cite{WangK10} proposes a fusion procedure trying to minimize some risks over the different modalities and a correlation measure.
While it is well-founded, it needs to define some appropriate functions and
is not completely fully adapted to the classifier fusion problem since it does not directly take into account the diversity between the  outputs of the classifiers.

We propose to study a new machine learning method, namely MinCq, introduced in \cite{MinCQ}. 
It proposes a quadratic program for learning a weighted majority vote over real-valued functions called voters (such as score functions of classifiers). The algorithm is based on the minimization of a generalization bound that takes into
account both the risk of committing an error and  the diversity of the voters, offering strong theoretical guarantees on
the learned majority vote.
In this article, our aim is to show the interest of this algorithm for
classifier fusion. We provide evidence that MinCq is able to
find good linear weightings but also very performing non-linear
combination with an extra kernel layer over the scores.
Since in multimedia retrieval, the performance measure is related to the rank of positive examples, we propose to extend MinCq to improve the Mean Average Precision. We base this extension on an additional order-preserving loss for verifying ranking pairwise constraints.

The paper is organized as follows. Section \ref{sec:mincq} deals with the theoretical framework of  MinCq. We extend MinCq as a late fusion method in Section \ref{sec:mincqpw}. Before concluding in Section \ref{sec:conclu}, we evaluate empirically the MinCq late fusion in Section \ref{sec:expe}.

%% file: notations.tex
\section{PAC-Bayesian MinCQ}
\label{sec:mincq}
In this section we present the algorithm MinCq of Laviolette {\it et al.} \cite{MinCQ} for learning a ${\cal Q}$-weighted majority vote of real-valued functions ({\it e.g.} classifier scores).
This method is based on the PAC-Bayes theory \cite{Mcallester99b}. 
We first recall the setting of MinCq.

We consider binary classification tasks over a {\it feature space} $X \subseteq \R^d$ of dimension $d$.
The {\it label space} is  $Y = \{-1,1\}$.
The training sample is 
$S = \{(\xbf_i,y_i)\}_{i=1}^{m}$ where each example $(\xbf_i,y_i)$ is drawn {\it i.i.d.} from a fixed --- but unknown --- probability distribution $\Dcal$ defined over $X \x  Y$.
We consider a space of real-valued voters $\Hcal$, such that $\forall h_i \in \Hcal,\ h_i : X \mapsto \R$.
Given a voter $h_i$, the predicted label of $\xbf \in  X$ is given by $\sign[h_i(\xbf)]$, where $\sign[a] = 1$ if $a \geq  0$ and $-1$ otherwise. 
Then, the learner aims at choosing a distribution $\Qcal$ over $\mathcal{H}$ ---  the weights $q_i$ --- leading to the {\it $\Qcal$-weighted majority vote} $B_{\Qcal}$ with the lowest risk. $B_{\Qcal}$ is defined by,
\begin{align*}
&B_{\Qcal}(\xbf) = \sign\left[H_{\Qcal}(\xbf)\right],\\&\textrm{with }H_{\Qcal}(\xbf)=\sum_{i=1}^{|\Hcal|}q_i h_i(\xbf).
\end{align*}
The associated true risk $R_{\Dcal}(B_{\Qcal})$ is defined as the probability that the majority vote  misclassifies an example drawn according to $\Dcal$,
\begin{align*}
R_{\Dcal}(B_{\Qcal}) = \PR_{(\xbf,y)\sim \mathcal{D}} \left(B_{\Qcal}(\xbf)\ne y\right).
\end{align*}
In the case of MinCq, $\Hcal$ has to be a finite {\it auto-complemented} family of $2n$ real-valued voters $\Hcal = \{h_1,\dots,h_{2n}\}$ such that,
\begin{align}
\label{eq:ac}
\forall \xbf\in X, \forall i\in\{1,\dots,n\},\ h_{i+n}(\xbf)\ =\ -h_{i}(\xbf).
\end{align}
Moreover, the algorithm considers  {\it quasi-uniform}  distributions $\Qcal$ over $\Hcal$, {\it i.e.}  the sum of the weight of a voter and its opposite is $\frac{1}{n}$, 
\begin{align}
\label{eq:qu}
\forall i\in\{1,\dots,n\},\ \Qcal(h_i)+\Qcal(h_{i+n})\ =\ q_i + q_{i+1}\ =\ \frac{1}{n}.
\end{align}
This constraint is not too restrictive since every distribution over $\Hcal$ can be represented by a quasi-uniform distribution \cite{MinCQ}.
The assumptions \eqref{eq:ac} and \eqref{eq:qu} are actually an elegant trick to avoid the use of a prior distribution over $\Hcal$ which is often required by usual PAC-Bayesian method \cite{Mcallester99b}, making the algorithm more easily applicable.

We now present the principle of the algorithm MinCq.
The core of MinCq  is  the minimization of the empirical version of a bound --- the $C$-Bound  --- over the risk of the $\Qcal$-weighted majority vote.
\begin{theorem}[$C$-Bound \cite{MinCQ}]
Given ${\cal H}=\{h_1,\dots, h_{2n}\}$ a class of $2n$ functions, for any weights $\{q_i\}_{i=1}^{2n}$, {\it i.e.}  distribution $\Wcal$, on ${\Hcal}$ and any distribution $\Dcal$ over $X \times  Y$, if $\ \E_{(\xbf,y)\sim {\cal D}}H_{\Wcal}(\xbf)  >  0$ then $R_\Dcal(B_\Qcal) \leq  C_\Qcal^\Dcal$ where,
\begin{align*}
C_\Qcal^\Dcal&= \frac{\Var_{(\xbf,y)\sim \Dcal} (y  H_{\Wcal}(\xbf))}{\E_{(\xbf,y)\sim \Dcal}(y H_{\Wcal}(\xbf))^2}  =  1 - \frac{( {\cal M}_\Qcal^\Dcal )^2  }{ {\cal M}_{\Qcal^2}^\Dcal},
\end{align*}
with ${\cal M}_\Qcal^\Dcal  =   \E_{(\xbf,y)\sim D} \sum_{i=1}^{2n} y q_ih_i(\xbf)$, and ${\cal M}_{\Qcal^2}^\Dcal   =  \E_{(\xbf,y)\sim\Dcal} \sum_{i=1}^{2n}\sum_{i'=1}^{2n} q_iq_{i'}h_i(\xbf)h_{i'}(\xbf)$  
are  respectively the first and the second moments of the $\Qcal$-margin: $y H_{\Wcal}(\xbf)$.
\end{theorem}
Following some generalization bounds, MinCq proposes to minimize the empirical version of the $C$-bound, $C_\Qcal^S  =  1 - \frac{(\Mcal_{\Wcal}^S)}{\Mcal_{\Wcal^2}^S}$, over a sample $S$.
The idea is to fix the empirical first moment $\Mcal_\Qcal^S$ to a margin $\mu > 0$ and to minimize the empirical second moment $\Mcal_{\Qcal^2}^S$ measuring the correlation of the voters.
This leads to minimize the bound and thus the risk of the majority vote by taking into account the diversity between the voters.
\begin{definition}[MinCq algorithm \cite{MinCQ}]
 Given a set ${\cal H} = \{h_1,\dots,h_{2n}\}$ of voters, a training set $S = \{(\xbf_j,y_j)\}_{j=1}^m$, and a margin $\mu > 0$, among all quasi-uniform distributions $\Qcal$ of empirical margin $\Mcal_{\Wcal}^{S}$  exactly equal to $\mu$, the MinCq algorithm consists in finding one that minimizes the empirical $\Mcal_{\Wcal^2}^S$.
\end{definition}
Due to the auto-complemented \eqref{eq:ac} and  quasi-uniformity \eqref{eq:qu} assumptions, the algorithm can be expressed as a quadratic program  \eqref{eq:mincq} by only considering the first $n$ voters $h_i \in \Hcal$. 
\begin{align}
\nonumber &\argmin_{\Qbf}\ \ \displaystyle \Qbf_S^t \Mbf_S \Qbf - \Abf_S^t \Qbf,\\
& \nonumber \textrm{s.t.}\ \  \displaystyle \mbf_S^t \Qbf =  \frac{\mu}{2}+ \frac{1}{2nm} \sum_{j=1}^m \sum_{i=1}^{n} y_jh_i(\xbf_j), \\
& \nonumber  \textrm{and}\ \  \forall i\in \{1,\dots,n\},\ \displaystyle0\leq q_i\leq\frac{1}{n}, \label{eq:mincq} \tag{\mbox{$MinCq$}}
\end{align}
where ${}^t$ denotes the transposed function, $\Qbf  =  (q_1,\dots,q_n)^t$ is the vector of the  first $n$  weights $q_i$, $\Mbf_S$ is the $n \times  n$ matrix formed by $\frac{1}{m}\sum_{j=1}^m h_i(\xbf_j)h_{i'}(\xbf_j)$ for $i$ and $i'$ in $\{1,\dots,n\}$, and,
\begin{align*} 
\mbf_S&=\bigg(\frac{1}{m}\sum_{j=1}^m y_jh_1(\xbf_j),\dots, \frac{1}{m}\sum_{j=1}^m y_jh_n(\xbf_j)\bigg)^t,\\
\Abf_S &= \bigg(\frac{1}{nm} \sum_{i=1}^n \sum_{j=1}^m h_1(\xbf_j)h_{i}(\xbf_j),\dots, \frac{1}{nm} \sum_{i=1}^n \sum_{j=1}^m h_n(\xbf_j)h_{i}(\xbf_j)\bigg)^t.
\end{align*}
Finally, the  $\Wcal$-weighted majority vote learned by MinCq is then 
\begin{align*}
&B_{\Wcal}(\xbf) = \sign[H_{\Wcal}(\xbf)],\\
&\textrm{ with } H_{\Wcal}(\xbf) =  \sum_{i=1}^{n} \left(2q_i  - \frac{1}{n}\right)h_i(\xbf).
\end{align*}

%% file: contrib.tex
\section{MinCq as a Late Fusion Method}
\label{sec:mincqpw}
PAC-Bayesian  MinCq has been proposed in the particular context of binary classification where the objective is to minimize the misclassification rate of the $\Qcal$-weighted majority vote by taking into account the diversity of the voters. 
From a  multimedia indexing standpoint, MinCq thus appears to be a natural way for late classifiers fusion to combine the predictions of classifiers separately trained from different modalities.

Concretely, given a training sample of size $2m$ we split it randomly into two subsets $S'$ and $S = \{(\xbf_j,y_j)\}_{j=1}^{m}$ of the same size.
Let $n$ be the number of modalities.
For each  modality $i$, we train a classifier $h_i$ from $S'$. 
Let $\Hcal = \{h_1,\dots,h_n,-h_i,\dots, -h_{n}\}$ be the set of the $n$ associated prediction functions and their opposites.
At this step, the fusion is achieved by MinCq: We learn from $S$ the $\Qcal$-weighted majority vote over $\Hcal$ with the lowest risk.
However, in many applications, such as multimedia document retrieval, people are interested in performance measures related to precision or recall. 
Since a low-error vote is not necessarily a good ranker, we propose an adaptation of MinCq  to improve the popular Mean Averaged Precision (MAP).

We first recall the definition of the MAP measured on $S$ for a given real-valued function $h$.
Let $S^+  = \{(\xbf_j,y_j)  :  (\xbf_j,y_j) \in  S \wedge y_j = 1\} = \{(\xbf_{j^+},1)\}_{j^+=1}^{m^+}$ be the set of the $m^+$ positive examples from $S$  and $S^-  = \{(\xbf_j,y_j)  :  (\xbf_j,y_j) \in  S \wedge  y_j = -1\} = \{(\xbf_{j^-},-1)\}_{j^-=1}^{m^-}$ the set of the $m^-$ negative examples from $S$ ($m^+  + m^-  = m$).
For evaluating the MAP, one ranks the examples in descending order of the scores.
The MAP of $h$ is,
\begin{align*}
MAP_S(h) = \frac{1}{|m^+|} \sum_{j:y_j=1} Prec@j,
\end{align*}
where $Prec@ j$ is the percentage of positive examples in the top $j$.
The intuition behind this definition is that we prefer positive examples with a score higher than negative ones. 
To achieve this goal, we propose to learn with {\it pairwise preference} \cite{Preference} on pairs of positive-negative instances.
Indeed, pairwise methods are known to be a good compromise between accuracy and more complex performance measure like MAP.
Especially, the notion of order-preserving pairwise loss was introduced in \cite{Zhang2004} in the context of multiclass classification.
Following this idea, Yue {\it et al.} \cite{YueFRJ07} have proposed a SVM-based method with a hinge-loss relaxation of a MAP-loss.
In our specific case of MinCq for multimedia fusion, we design an order-preserving pairwise loss for correctly ranking the positive examples.
Actually, for each pair  $(\xbf_{j^+} ,\xbf_{j^-} ) \in  S^+ \times  S^-$, we want:
$H_{\Wcal}(\xbf_{j^+})  >  H_{\Wcal}(\xbf_{j^-}) \Leftrightarrow H_{\Wcal}(\xbf_{j^-})  -  H_{\Wcal}(\xbf_{j^+})  <  0.$
This can be forced by minimizing (according to the weights $q_i$) the following hinge-loss relaxation of the previous equation,
\begin{align}
\label{eq:PW}
 \frac{1}{m^+m^-}\sum_{j^+=1}^{m^+} \sum_{j^-=1}^{m^-} \left[H_{\Wcal}(\xbf_{j^-}) - H_{\Wcal}(\xbf_{j^+})\right]_+,
\end{align}
where $[a]_+=\max(a,0)$ is the hinge-loss.
In the setting of MinCq,  with $\Hcal$ auto-complemented (Eq.\eqref{eq:ac}) and $\Wcal$ quasi-uniform (Eq.\eqref{eq:qu}), we reduce the term \eqref{eq:PW} 
to,
\begin{align}
\label{eq:PWsimp}
\frac{1}{m^+m^-}  \sum_{j^+=1}^{m^+}  \sum_{j^-=1}^{m^-}  \left[\sum_{i=1}^n  \left( 2q_i - \frac{1}{n} \right)   \left(h_i(\xbf_{j^-})  -  h_i(\xbf_{j^+}) \right) \right]_+    .
\end{align}
To deal with the hinge-loss of \eqref{eq:PWsimp}, we consider  $m^+ \times m^-$ additional {\it slack variables $\boldsymbol{\xi}_{S^+\x S^-}  =  (\xi_{j^+j^-})_{1\leq j^+ \leq m^+,1\leq j^-\leq m^-} $} weighted by a parameter $\beta  >  0 $. 
We make a little abuse of notation to highlight the difference with \eqref{eq:mincq}: Since $\boldsymbol{\xi}_{S^+\x S^-}$ appear only in the linear term, we simply add \eqref{eq:PWsimp} after the \eqref{eq:mincq} formulation.
We obtain the quadratic program \eqref{eq:mincqpw},
\begin{align}
\nonumber &\argmin_{\Qbf,\boldsymbol{\xi_{S^+\x S^-}}}\ \ \displaystyle \Qbf_S^t \Mbf_S \Qbf - \Abf_S^t \Qbf +  \beta\ {\bf Id}^t\boldsymbol{\xi}_{S^+\x S^-},\\
\nonumber & \textrm{s.t.}\ \  \displaystyle \mbf_S^t \Qbf =  \frac{\mu}{2}+ \frac{1}{2nm} \sum_{j=1}^m \sum_{i=1}^{n} y_jh_i(\xbf_j),\\
\nonumber &\qquad \forall j^+\in\{1,\dots,m^+\},\forall j^-\in\{1,\dots,m^-\},\ \xi_{j^+j^-} \geq 0,\\
\nonumber & \qquad\displaystyle \xi_{j^+j^-} \geq \frac{1}{m^+m^-}\sum_{i=1}^n  \left( 2q_i - \frac{1}{n} \right)   \left(h_i(\xbf_{j^-})  -  h_i(\xbf_{j^+}) \right),\\
&  \textrm{and}\ \  \forall i\in \{1,\dots,n\},\ \displaystyle0\leq q_i\leq\frac{1}{n},  \label{eq:mincqpw} \tag{\mbox{$MinCq_{PW}$}}
\end{align}
where ${\bf Id}$ is the unit vector of size $m^{+} \times m^{-}$.
However, one drawback of this method is the incorporation of a quadratic number of additive variables ($m^{+} \times m^{-}$) which makes the problem harder to solve.
To overcome this problem, we propose to relax the constraints by considering the average score of the negative examples: We force the positive examples to be higher than the average negative scores.
This leads us to the following alternative problem \eqref{eq:mincqpwav} with only $m^+$ additional variables.
\begin{align}
\nonumber &\argmin_{\Qbf,\boldsymbol{\xi_{S^+}}}\ \ \displaystyle \Qbf_S^t \Mbf_S \Qbf - \Abf_S^t \Qbf +  \beta\ {\bf Id}^t\boldsymbol{\xi}_{S^+},\\
\nonumber& \textrm{s.t.}\ \  \displaystyle \mbf_S^t \Qbf =  \frac{\mu}{2}+ \frac{1}{2nm} \sum_{j=1}^m \sum_{i=1}^{n} y_jh_i(\xbf_j),\\
\nonumber &\qquad \forall j^+\in\{1,\dots,m^+\},\ \xi_{j^+} \geq 0,\\\
\nonumber & \qquad\displaystyle  \xi_{j^+} \geq \frac{1}{m^+m^-} \sum_{j^-=1}^{m^-}  \sum_{i=1}^n   \left( 2q_i - \frac{1}{n} \right)   \left(h_i(\xbf_{j^-})  -  h_i(\xbf_{j^+}) \right),\\
&  \textrm{and}\ \  \forall i\in \{1,\dots,n\},\ \displaystyle0\leq q_i\leq\frac{1}{n},  \label{eq:mincqpwav} \tag{\mbox{$MinCq_{PWav}$}}
\end{align}
where ${\bf Id}$ is the unit vector of size $m^+$.

 Note that the two approaches still respect the framework of the original MinCq. We simply regularize the search of the weights for  a $\Qcal$-weighted majority vote leading to an higher MAP. 

Finally, for tuning the hyperparameters ($\mu$, $\beta$) we use a cross-validation process (CV).
Instead of selecting the parameters leading to the lowest risk, we select the ones leading to the best MAP.

%% file: experiments.tex
\section{Experiments}
\label{sec:expe}

\input{tableau_res}

In this section, we show empirically the interest of MinCq, and our extension, as a late fusion method with stacking (implemented with MOSEK solver).
We experiment the MinCq-based approaches on the PascalVOC'07 benchmark \cite{voc2007}, where the goal is a list of $20$ visual concepts to identify in images.
The corpus is constituted of $5000$ training and $5000$ test images.
In general, the ratio between positive and negative examples is less than $10\%$.
For each concept, we generate a training sample constituted of all the training positive examples and negative examples independently drawn such that the positive ratio is $1/3$.
We keep the original test set.

Our objective is not to provide the best results on this benchmark but rather to evaluate if the MinCq-based methods could be helpful for the late fusion step in multimedia indexing.
To do so, we split the training sample into two subsets, $S'$ and $S$, of the same size.
We consider $9$ different visual features: $1$ SIFT, $1$ LBP, $1$  Percepts, $2$ HOG, $2$ Local Color Histograms and $2$ Color Moments. 
Then, we train from $S'$ a SVM-classifier for each visual feature (with the LibSVM library \cite{libsvm} and a rbf kernel with parameters tuned by CV).
The final classifier fusion is learned from $S$.

In a first series of experiments, the set of voters $\Hcal$ is constituted by the  $9$ SVM-classifiers (MinCq also considers the opposites).
We compare the $3$ linear MinCq methods \eqref{eq:mincq}, \eqref{eq:mincqpw}, \eqref{eq:mincqpwav} to the following $4$ baseline fusion approaches.
\begin{itemize}
\item The best classifier of $\Hcal$: $$h_{best}  =  \argmax_{h_i\in\Hcal}  MAP_{S}(h_i).$$ 
\item The one with the highest margin: $$best(\xbf)  =  \argmax_{h_i\in\Hcal}  |h_i(\xbf)|.$$ 
\item The sum of the classifiers (unweighted vote): $$\Sigma(\xbf)  = \sum_{h_i\in\Hcal} h_i(\xbf).$$ 
\item The MAP-weighted vote: $$\Sigma_{MAP}(\xbf)  =  \sum_{h_i\in\Hcal}   \frac{MAP_{S}(h_i)}{\sum_{h_{i'} \in\Hcal}MAP_{S} (h_{i'} )}h_i(\xbf).$$
\end{itemize}

In a second series, we propose to introduce non-linear information with a rbf kernel layer.
We represent each example by the vector of its scores of the $9$ SVM-classifiers, $\Hcal$ being the set of kernels over the sample $S$: Each  $\xbf \in S$ is seen as a voter $k(\cdot,\xbf)$.
We then compare our method to stacking with SVM tuned by CV (SVM$^{rbf}$).
Note that we do not report the results of \eqref{eq:mincqpw} in this context, because the computational cost is much higher and the performance is lower. The full pairwise version implies too many variables which may penalize the resolution of \eqref{eq:mincqpw}.

In either case, the hyperparameters of MinCq-based methods are tuned with a grid search by a $5$-folds CV. 
The MAP-performances are reported on Table \ref{tab:res}, we can make the following remarks.
\begin{itemize}
\item On the right, for the first experiments, we clearly see that the linear MinCq-based algorithms outperform on average the linear baselines. 
At least one MinCq-based method produces the highest MAP, except for ``boat'' for which $h_{best}$ is the best.
We note that the order-preserving hinge-loss is not really helpful: The classical \eqref{eq:mincq}  shows the best MAP.
In fact, this can be explained by the limited number of voters.
\item  On the left, with a kernel layer, at least one MinCq-based method achieves the highest MAP and for $17/20$ both are better than SVM. Moreover,  $MinCq_{PWav}^{rbf} $ with the averaged pairwise preference is the best for $17$ concepts, showing the order-preserving loss is a good compromise between improving the MAP and keeping a reasonable computational cost.
\item Globally, kernel-based MinCq methods outperform the other methods.
Moreover,  at least one MinCq-based approach is the best for each concept showing PAC-Bayesian MinCq is a good alternative for late classifiers fusion.
\end{itemize}

%% file: tableau_res.tex
\begin{table*}
\caption{MAP obtained on the PascalVOC'07 test sample. On the left, experiments with rbf kernel layer. On the right, without.}
\label{tab:res}
\centering
\scriptsize
\begin{tabular}{|c||c|c|c||c|c|c|c|c|c|c|}
\hline
concept &	 $\!MinCq_{PWav}^{rbf}\!\!$  &	 $\!MinCq^{rbf}\!$  &	 SVM$^{rbf}\!$  &	 $\!MinCq_{PWav}\!\!$  &	 $\!MinCq_{PW}\!$  &	 $MinCq$  &	 $\Sigma$  &	 $\Sigma_{MAP}$  &	 $best$  &	 $h_{best}$ \\			
\hline			
\hline																	
aeroplane  &	 $\boldsymbol{\it 0.513}	$  &	 $ \boldsymbol{\it 0.513}	$  &	 $ 0.497	$  &	 $ 0.487	$  &	 $ 0.486	$  &	 $ \mathbf{0.526}	$  &	 $ 0.460	$  &	 $ 0.241	$  &	 $ 0.287	$  &	 $ 0.382$ \\										
\hline	
bicycle	 &	 $ \mathbf{0.273} $  &	 $ 0.219	$  &	 $ 0.232	$  &	 $ 0.195	$  &	 $ 0.204	$  &	 $ \boldsymbol{\it 0.221}	$  &	 $ 0.077	$  &	 $ 0.086	$  &	 $ 0.051	$  &	 $ 0.121$ \\	
\hline	
bird	& 	 $ 0.2659	$  &	 $\mathbf{0.264}	$  &	 $ 0.196	$  &	 $ 0.169	$  &	 $ 0.137	$  &	 $\boldsymbol{\it 0.204}	$  &	 $ 0.110	$  &	 $ 0.093	$  &	 $ 0.113	$  &	 $ 0.123$ \\							
\hline	
boat	& 	 $ \mathbf{0.267}	$  &	 $ 0.242	$  &	 $ 0.240	$  &	 $ 0.1593	$  &	 $ 0.154	$  &	 $ 0.159	$  &	 $ 0.206	$  &	 $ 0.132	$  &	 $ 0.079	$  &	 $\boldsymbol{\it 0.258}$ \\	
\hline	
bottle	& 	 $ \boldsymbol{\it 0.103}	$  &	 $ 0.099	$  &	 $ 0.042	$  &	 $ 0.112	$  &	 $ \mathbf{0.126}	$  &	 $0.118	$  &	 $ 0.023	$  &	 $ 0.025	$  &	 $ 0.017	$  &	 $ 0.066$ \\	
\hline
bus	& 	 $\mathbf{0.261}	$  &	 $ \mathbf{0.261}	$  &	 $ 0.212	$  &	 $\boldsymbol{\it 0.167}	$  &	 $ 0.166	$  &	 $ 0.168	$  &	 $ 0.161	$  &	 $ 0.098	$  &	 $ 0.089	$  &	 $ 0.116$ \\
\hline
car	& 	 $ \mathbf{0.530}	$  &	 $ \mathbf{0.530}	$  &	 $ 0.399	$  &	 $\boldsymbol{\it 0.521}	$  &	 $ 0.465	$  &	 $ 0.495	$  &	 $ 0.227	$  &	 $ 0.161	$  &	 $ 0.208	$  &	 $ 0.214$ \\	
\hline			
cat	& 	 $ \mathbf{ 0.253}	$  &	 $ 0.245	$  &	 $ 0.160	$  &	 $\boldsymbol{\it 0.230}	$  &	 $ 0.219	$  &	 $ 0.220	$  &	 $ 0.074	$  &	 $ 0.075	$  &	 $ 0.065	$  &	 $ 0.116$ \\		
\hline	
chair	 &	 $\mathbf{0.397}	$  &	 $\mathbf{ 0.397}	$  &	 $ 0.312	$  &	 $ \boldsymbol{\it 0.257}	$  &	 $ 0.193	$  &	 $ 0.230	$  &	 $ 0.242	$  &	 $ 0.129	$  &	 $ 0.178	$  &	 $ 0.227$ \\	
\hline	
cow	 &	 $ 0.158	$  &	 $\mathbf{0.177}	$  &	 $ 0.117	$  &	 $ 0.102	$  &	 $ 0.101	$  &	 $\boldsymbol{\it 0.118}	$  &	 $ 0.078	$  &	 $ 0.068	$  &	 $ 0.06	$  &	 $ 0.101$ \\			
\hline	
diningtable	 &	 $\mathbf{0.263}	$  &	 $ 0.227	$  &	 $ 0.245	$  &	 $ 0.118	$  &	 $ 0.131	$  &	 $ 0.149	$  &	 $\boldsymbol{\it 0.153}	$  &	 $ 0.091	$  &	 $ 0.093	$  &	 $ 0.124$ \\
\hline	
dog	& 	 $\mathbf{ 0.261}	$  &	 $ 0.179	$  &	 $ 0.152	$  &	 $ \boldsymbol{\it 0.260}	$  &	 $ 0.259	$  &	 $ 0.253	$  &	 $ 0.004	$  &	 $ 0.064	$  &	 $ 0.028	$  &	 $ 0.126$ \\
\hline
horse	 &	 $\mathbf{ 0.495}	$  &	 $ 0.4504	$  &	 $ 0.437	$  &	 $ 0.3011	$  &	 $ 0.259	$  &	 $ 0.303	$  &	 $\boldsymbol{\it 0.364}	$  &	 $ 0.195	$  &	 $ 0.141	$  &	 $ 0.221$ \\
\hline	
motorbike  &	 $\mathbf{ 0.295}	$  &	 $ 0.284	$  &	 $ 0.207	$  &	 $ 0.1412	$  &	 $ 0.113	$  &	 $ 0.162	$  &	 $ \boldsymbol{\it 0.193}$  &	 $ 0.115	$  &	 $ 0.076	$  &	 $ 0.130$ \\
\hline
person	& 	 $ \mathbf{0.630}	$  &	 $ 0.614	$  &	 $ 0.237	$  &	 $ \boldsymbol{\it 0.624}	$  &	 $ 0.617	$  &	 $ 0.604	$  &	 $ 0.001	$  &	 $ 0.053	$  &	 $ 0.037	$  &	 $ 0.246$ \\	
\hline	
pottedplant	 &	 $ 0.102	$  &	 $\mathbf{ 0.116}	$  &	 $ 0.065	$  &	 $\boldsymbol{\it 0.067}	$  &	 $ 0.061	$  &	 $ 0.061	$  &	 $ 0.057	$  &	 $ 0.04	$  &	 $ 0.046	$  &	 $ 0.073$ \\	
\hline			
sheep	 &	 $ \mathbf{0.184}	$  &	 $ 0.175	$  &	 $ 0.144	$  &	 $ 0.0666	$  &	 $\boldsymbol{\it 0.096}	$  &	 $ 0.0695	$  &	 $ 0.128	$  &	 $ 0.062	$  &	 $ 0.064	$  &	 $ 0.083$ \\	
\hline		
sofa	 &	 $\mathbf{0.246}	$  &	 $ 0.211	$  &	 $ 0.162	$  &	 $ 0.204	$  &	 $ \boldsymbol{\it 0.208}	$  &	 $ 0.201	$  &	 $ 0.137	$  &	 $ 0.087	$  &	 $ 0.108	$  &	 $ 0.147$ \\	
\hline
train	 &	 $ \mathbf{0.399}	$  &	 $ 0.385	$  &	 $ 0.397	$  &	 $ 0.331	$  &	 $ 0.332	$  &	 $ \boldsymbol{\it 0.335}	$  &	 $ 0.314	$  &	 $ 0.164	$  &	 $ 0.197	$  &	 $ 0.248$ \\							
\hline			
tvmonitor  &$\boldsymbol{\it 0.272}	$  &	 $ 0.257	$  &	 $ 0.230	$  &	 $\mathbf{0.281}	$  &	 $ \mathbf{0.281}	$  &	 $ 0.256	$  &	 $ 0.015	$  &	 $ 0.102	$  &	 $ 0.069	$  &	 $ 0.171$ \\										
\hline	
\hline	
Average	 &$ \mathbf{0.301}	$  &	 $ 0.292	$  &	 $ 0.234$ &	 $ 0.240	$  &	 $ 0.231	$  &	 $ \boldsymbol{\it  0.243}	$  &	 $ 0.151	$  &	 $ 0.104	$  &	 $ 0.100	$  &	 $ 0.165$ \\	
\hline	
\end{tabular}
\end{table*}

%% file: conclusion.tex
\section{Conclusion}
\label{sec:conclu}
We propose in this paper to make use of a well-founded learning quadratic program called MinCq as a novel multimedia late fusion method.
PAC-Bayesian MinCq  was originally developed for binary classification and aims at minimizing the error rate of the weighted majority vote by considering the diversity of the voters \cite{MinCQ}.
In the context of multimedia indexing, we claim that MinCq thus appears naturally appropriate for late classifier fusion in order to combine the predictions of classifiers trained from different modalities.
Our experiments show that MinCq is a very competitive alternative for classifier fusion. 
Moreover, the incorporation of average order-preserving constraints is sometimes able to improve the MAP-performance measure.
Beyond these results, such PAC-Bayesian methods open the door  to define other theoretically well-founded frameworks to design new algorithms in many multimedia tasks such as multi-modality indexing, multi-label classification, ranking, etc.